\documentclass[]{fairmeta}

\usepackage{amsmath}
\usepackage{amssymb}
\usepackage{wrapfig}
\usepackage{enumitem}
\usepackage{arydshln}

\providecommand{\ourssft}{\texttt{MobileLLM-R2-SFT}}
\providecommand{\cttok}[1]{\texttt{\detokenize{#1}}}

\usepackage[most]{tcolorbox}
\newtcblisting{protocolbox}{
  listing only,
  listing options={
    basicstyle=\ttfamily\scriptsize,
    breaklines=true,
    columns=fullflexible,
    keepspaces=true
  },
  colback=gray!3,
  colframe=gray!45,
  boxrule=0.4pt,
  arc=1pt,
  left=4pt, right=4pt, top=4pt, bottom=4pt
}

\newcommand{\ours}{\texttt{MobileLLM-R2}}

\title{$\ours{}$: Advancing On-Device Language Models from Reasoning to Agents}

\author[1,*]{Changsheng Zhao}
\author[1,*,\S]{Ernie Chang}
\author[1,*,\dagger]{Zechun Liu}
\author[1]{Dylan R. Ashley}
\author[1]{Matteo Paltenghi}
\author[1]{Vikas Chandra}
\author[1]{Yangyang Shi}
\author[1]{Raghuraman Krishnamoorthi}

\affiliation[1]{Meta AI}

\contribution[*]{Equal contribution, authors listed in alphabetical order \\}
\contribution[\S]{Led overall data curation efforts \\}
\contribution[\dagger]{Corresponding author; Project lead}

\correspondence{Zechun Liu at \email{zechunliu@meta.com}}
\metadata[Author emails]{\email{\{cszhao, erniecyc, zechunliu\}@meta.com}}
\metadata[Project page]{https://facebookresearch.github.io/MobileLLM/R2/}
\metadata[HuggingFace]{https://huggingface.co/facebook/MobileLLM-R2-2.5B}

\abstract{
Recent advances in agentic models are expanding the role of language models from reasoning and responding to interacting with applications and acting on behalf of users. Bringing this transition to on-device intelligence requires developing reliable agentic capabilities within stringent memory and computational budgets. Existing work, however, either develops proprietary construction recipes for large-scale models or optimizes post-training in isolation, leaving the end-to-end training of small agentic models poorly understood. We show that downstream agentic performance is strongly conditioned on the capabilities and behavioral support established during pre-training; post-training sharpens what the base model can already express, but does not reliably recover what it cannot. We recursively optimize pre-training and mid-training data mixtures through influence-based selection and performance feedback, introducing synthetic cross-domain bridge data to mitigate capability competition. We identify sampling lift as an early indicator of post-training potential and develop a readiness predictor to anticipate diminishing returns from pre-training. During post-training, we demonstrate that joint supervision accommodates the distinct reasoning patterns of single-turn deliberation and multi-turn agentic interaction. We further identify structural and trajectory distribution alignment during SFT as critical to effective on-policy distillation, motivating chat protocol adaptation and teacher rewriting from both initial prompts and student-induced intermediate states. Together, these findings establish an end-to-end training recipe for compact agentic models. Trained from scratch with only 1.2T pre-training tokens, our 2.5B model achieves the best performance on AIME’24, LiveCodeBench, SWE-Bench Verified, Terminal-Bench, and Aider Polyglot among evaluated models up to 7.3B parameters, with a substantial 15.3-percentage-point lead on SWE-Bench Verified. Our training pipeline is built predominantly on publicly available data sources, as documented in this paper. We have also released our model checkpoints to facilitate reproducibility and further research.
}

\begin{document}

\maketitle

\section{Introduction}

Recent progress in agentic AI, where models plan, call tools, and act autonomously~\citep{yao2022react,schick2023toolformer,gao2023pal}, has been driven largely by frontier-scale models trained by a small number of well-resourced labs (e.g., Claude~\citep{anthropic2024claude35sonnet} and Kimi~K2~\citep{team2025kimi}). This concentration means that architectural and training innovations for agentic capability rarely reach the broader community: most teams that need an agentic model for a specific, in-domain task are limited to prompting or lightly fine-tuning an existing frontier or open-weight model, rather than participating in how such capability is built from the ground up. Yet a growing number of agentic systems are shifting toward small, specialized models (e.g., Octopus~v2~\citep{chen2024octopus}, TinyAgent~\citep{erdogan2024tinyagent}, Hammer~\citep{lin2024hammer}) precisely because narrow, tool-heavy workloads do not require frontier-scale general intelligence; they require reliability, low latency, and low cost on a bounded task distribution. Meanwhile, mobile device manufacturers are also exploring a new interaction paradigm in which language models serve as an intelligent interface between users and applications, translating natural-language intentions into coordinated actions~\citep{apple_app_intents,google_gemini_apps,honor_magicos8}. These developments raise a fundamental question: \textit{how can reliable agentic capabilities be developed from scratch within the stringent model capacity and computational constraints of on-device deployment}? To investigate this question, we focus on a deployment-relevant model size of 2.5B parameters, with our model requiring only 3.0 GiB of inference memory at a 32K-token context under weight and KV-cache quantization on an Apple M4 Max, suggesting memory feasibility on commodity mobile devices.

\begin{figure}[tb!]
    \centering
    \includegraphics[width=\textwidth]{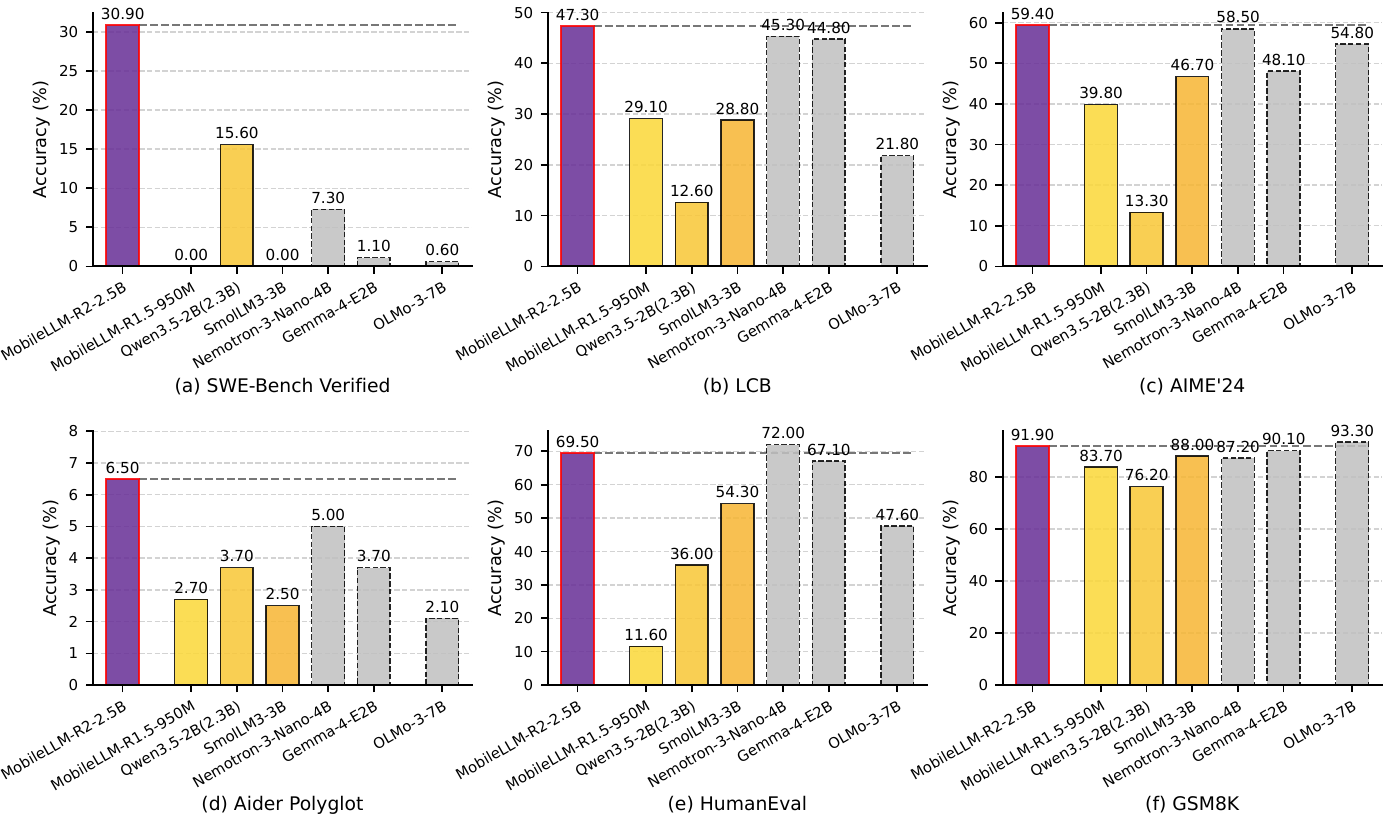}
    \vspace{-1em}
    \caption{\small Performance comparison across reasoning, coding, and agentic benchmarks. With only 2.5B parameters and 1.2T pre-training tokens, MobileLLM-R2 outperforms OLMo-3-7B-Instruct (7.3B parameters, 5.9T pre-training tokens) on five of the six benchmarks shown, spanning mathematical reasoning, coding, and agentic tasks. Notably, MobileLLM-R2 achieves the highest accuracy on AIME’24, LiveCodeBench, SWE-Bench Verified, and Aider Polyglot among all evaluated models, surpassing the strongest baseline by 15.3 percentage points on SWE-Bench Verified. Full results are reported in Table~\ref{tab:overall_acc}.}
    \label{fig:overall_accuracy}
\end{figure}

Existing recipes for small agentic models formulate agentic capability acquisition primarily as a post-training problem: they curate supervised fine-tuning or distillation data over tool-calling trajectories while treating pre-training as a fixed, off-the-shelf ingredient (see, e.g., ToolLLM~\citep{qin2024toolllm}, APIGen~\citep{liu2024apigen}, ToolACE~\citep{liu2025toolace}, APIGen-MT~\citep{prabhakar2026apigen}). This practice implicitly assumes that any reasonably capable pre-trained model can reach a similar agentic endpoint when paired with a sufficiently strong post-training recipe. Our findings challenge this assumption. Through interventions across the training pipeline, we find that downstream agentic trainability is path-dependent: the composition and serialization of pre-training data substantially affect both the data efficiency and the attainable performance of subsequent agentic post-training. Introducing the same capability only during post-training does not close this gap within the data regime we study.

Establishing a balanced capability foundation from scratch is particularly challenging for small models, where limited capacity makes data relevance and cross-domain capability interactions especially consequential. We recursively optimize pre-training and mid-training data mixtures through influence-based data selection and model performance feedback, and introduce synthetic cross-domain bridge data to mitigate capability competition rather than simply reweighting competing domains. Beyond data composition, we investigate which pre-training signals predict downstream learnability. We identify sampling lift, $\Delta_k = \operatorname{pass}@k - \operatorname{pass}@1$~\citep{brown2024large,chen2021evaluating}, as a complementary signal to zero-shot accuracy that captures solvable capability beyond single-sample performance. We further find that pre-trained capabilities differ substantially in their predictive power for downstream agentic performance, motivating a readiness predictor that guides checkpoint selection and identifies diminishing returns during pre-training.

Building on this foundation, we jointly train on long-horizon reasoning and multi-turn agentic trajectories during SFT, enabling a compact model to accommodate distinct reasoning patterns. We further find that effective on-policy distillation (OPD) requires both structural and behavioral alignment during SFT, as interaction protocols and learned behavioral patterns persist in student rollouts. We establish this alignment through chat template adaptation and teacher rewriting of SFT solutions. For multi-turn agentic tasks, where student actions alter subsequent interaction states, we generalize teacher rewriting from initial prompts to student-induced intermediate states, allowing the teacher to complete the remaining interaction.

We validate these findings end-to-end across both agentic and general reasoning capabilities. We use SWE-Bench Verified as our primary agentic benchmark, as it requires multi-turn planning, tool invocation, error recovery, and long-horizon state tracking, which are foundational capabilities for application-level agents. With only 1.2T pre-training tokens, roughly 30$\times$ fewer than Qwen3-1.7B~\citep{yang2025qwen3}, our 2.5B-parameter model achieves the highest accuracy on AIME'24 (59.4\%), LiveCodeBench (47.3\%), and SWE-Bench Verified (30.9\%) among evaluated models up to 7.3B parameters, with a substantial 15.3-percentage-point lead on SWE-Bench Verified, as shown in Figure~\ref{fig:overall_accuracy}. Our training pipeline relies predominantly on publicly available data sources, with detailed data compositions provided in Tables~\ref{tab:pretraining-data-sources}, \ref{table:mid-train-data}, and \ref{tab:post-training_data}.

\begin{figure}[tb!]
    \centering
    \includegraphics[width=0.85\textwidth]{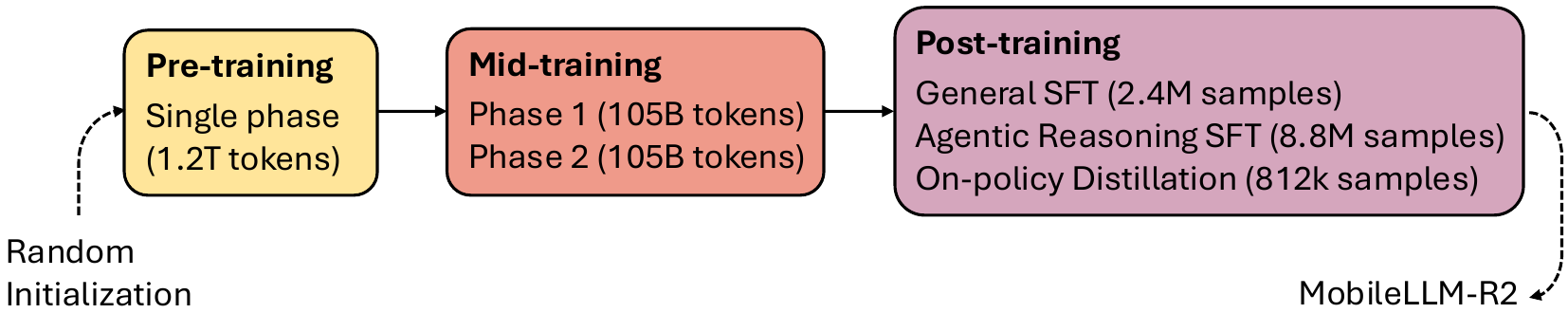}
    \caption{\small Overall training pipeline of $\ours{}$.}
    \label{fig:overall_pipeline}
\end{figure}

\section{Recursive Optimization of Pre-training Data Mixtures}
\label{sec:pretraining}

\begin{figure*}[t]
  \centering
  \includegraphics[width=\textwidth]{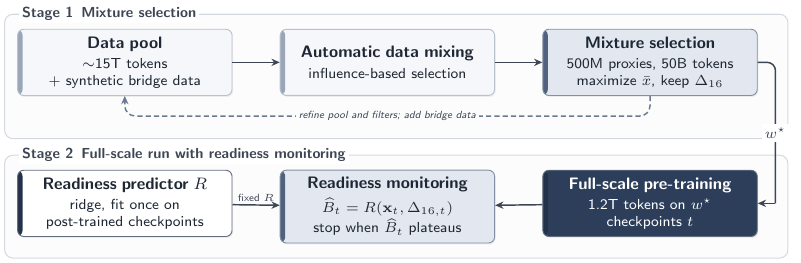}
  \caption{\small Data curation pipeline.
  \textbf{Stage 1 (mixture selection):} automatic data mixing derives candidate mixtures over the data pool, which includes synthetic bridge data, by influence-based sample selection in simulated runs; candidates are ranked by zero-shot scores averaged within and then across capability categories, together with sampling lift, and the best becomes the final mixture $w^\star$.
  \textbf{Stage 2 (full-scale run):} pre-training on $w^\star$ is monitored by a readiness predictor, fit once on post-trained checkpoints, to detect diminishing returns.}
  \label{fig:rsi-data-pipeline}
\end{figure*}

As illustrated in Figure~\ref{fig:rsi-data-pipeline}, we optimize the data mixtures for both pre-training and mid-training through a recursive, performance-driven pipeline that continuously refines data proportions based on model performance, rather than relying on manually crafted mixing ratios.
Section~\ref{sec:data-pool} describes the data sources and Section~\ref{sec:data-processing} how they are processed.
Section~\ref{sec:data-design} describes how the pre-training and mid-training mixtures are constructed around target capabilities, and Section~\ref{sec:bridge} describes the synthetic cross-domain \emph{bridge} data.
Section~\ref{sec:readiness} describes the readiness predictor that monitors the full-scale run.

\subsection{Data Sources}
\label{sec:data-pool}

Building on the capability-targeted datasets of MobileLLM-R1~\citep{zhao2026mobilellm}, we expand our data pool to over 15T raw tokens from 194 sources, from which we construct the pre-training data mixture through our recursive data curation pipeline.
It consists primarily of publicly released datasets:
web text from filtered Common Crawl derivatives \citep{li2024datacomp,penedo2024fineweb,soldaini2024dolma},
code from permissively licensed repositories \citep{lozhkov2024starcoder},
mathematics from web-mined and curated math corpora \citep{paster2024openwebmath,allal2025smollm2},
and additional multi-source collections \citep{soldaini2024dolma}.
The remainder comes from capability-targeted sources spanning knowledge, commonsense, reasoning, science, mathematics, and code, together with the synthetic data described in \S\ref{sec:bridge}.

Because the pool is larger than the pre-training budget, some sources are subsampled and others are repeated. The resulting data mixture, including the selected sources and their corresponding sampling weights, is detailed in Table~\ref{tab:pretraining-data-sources}.
We cap repetition at {4} epochs per source, following evidence that returns from repeated data diminish quickly beyond a few epochs~\citep{muennighoff2023scaling}.

\subsection{Data Processing}
\label{sec:data-processing}

\paragraph{Deduplication.}
Several sources overlap.
Dolma~3 CC~\citep{olmo2025olmo3}, DCLM~\citep{li2024datacomp}, and FineWeb-Edu~\citep{penedo2024fineweb} are all derived from Common Crawl~\citep{commoncrawl}; ProofPile ArXiv~\citep{azerbayev2024llemma} and ArXiv~\citep{weber2024redpajama} both contain arXiv papers; and Wikipedia~\citep{wikidump} appears in two sources.
We therefore deduplicate across sources as well as within them, since duplicated training data degrades language models and inflates memorization~\citep{lee2022deduplicating}.
We first remove exact duplicates by content hash, and then remove near-duplicates by semantic deduplication~\citep{abbas2023semdedup,tirumala2023d4} using {embeddings from an off-the-shelf sentence encoder~\citep{reimers2019sentence} with a cosine-similarity threshold of 0.92}.
Deduplication removes {11.6}\% of pool tokens, most of it between the Common Crawl derivatives.
Mid-training sources are additionally deduplicated against the pre-training pool.

\paragraph{Quality filtering.}
For each source we retain the upstream quality filters including FineWeb-Edu filters and other quality filters automatically built from source domain set based on their influence scores using trained proxy models~\cite{automixer}.
Additionally, we apply {a language-identification filter (English, fastText score $\geq 0.65$) and a document-length filter (50 to 100{,}000 words)}.
We do not train a separate quality classifier.
Instead, the influence-based selection of \S\ref{sec:data-design} acts as a relevance signal: it scores each sample by its estimated effect on the target capabilities rather than by a generic notion of quality.

\paragraph{Decontamination.}
The benchmarks used to rank candidate mixtures also feed the model readiness predictor in \S\ref{sec:readiness}, and influence-based selection pulls data toward the target set, so contamination would bias both the mixture and the predictor.
We decontaminate all pre-training, mid-training, and synthetic data against the 20 benchmarks of Table~\ref{tab:readiness-benchmarks} and all evaluation sets, removing any document that shares a {13}-gram with a benchmark item; this removes {2.3M} documents ({0.4}\% of pool tokens).
The target set used for automatic mixing shares no items with any evaluation set.

\begin{table}[t]
\small
  \centering
  \caption{Zero-shot benchmarks used to compare candidate mixtures and as inputs to the readiness predictor.}
  \vspace{-0.5em}
  \setlength{\tabcolsep}{1.5mm}
  \renewcommand{\arraystretch}{1.02}
  \label{tab:readiness-benchmarks}
  \begin{tabular}{@{}l l c@{}}
    \toprule
    Category & Benchmarks & \# \\
    \midrule
    Knowledge   & MMLU, MMLU-Redux, MMLU-Pro, BoolQ          & 4 \\
    Science     & ARC-Challenge, ARC-Easy, GPQA, SuperGPQA   & 4 \\
    Commonsense & HellaSwag, PIQA, SIQA, WinoGrande          & 4 \\
    Mathematics & GSM8K, MATH                                & 2 \\
    Reasoning   & BBH, BBEH                                  & 2 \\
    Code        & HumanEval+, MBPP, CRUXEval, MultiPL-E      & 4 \\
    \midrule
    Total       &                                            & 20 \\
    \bottomrule
  \end{tabular}
\end{table}

\subsection{Data Mixture}
\label{sec:data-design}
\label{sec:auto-mixing}

Our data design supports two capability families within a fixed pre-training token budget: \emph{agentic} reasoning, such as multi-turn tool use and instruction following, and \emph{long-horizon deliberative} reasoning, such as mathematics and code.
We therefore optimize the data mixture against both families jointly rather than treating them independently.

\paragraph{Automated data mixing.}
Mixture weights are derived by an extension of coverage data-mixing that leverages sample influence as utility~\citep{automixer} where we cluster random sampled domain data across all data domains consisting math, code, and web data into 64 clusters and score each candidate sample with an $n$-gram influence estimator, which measures its overlap in $n$-gram statistics with a target cluster.
A stochastic accept/reject step then selects samples according to their influence scores, with the selection budget ranging from 50B to 100B tokens balanced across clusters so that no single cluster dominates.
Each source's weight is its share of the accepted samples.
These simulated runs involve sampling only, with no model training, and repeat for at most five rounds until the weights change by less than a tolerance $\epsilon$.
One full mixing run costs {about 2{,}000 CPU-hours}, negligible relative to a single proxy training run.

\paragraph{Recursive refinement.}
Each resulting mixture is a candidate that we train at small scale with 50B tokens and evaluate; the scores inform the source pool, filtering thresholds, and sampling ratios for the next round, and we stop when scores plateau.
When a capability remains weak across rounds despite reweighting, we generate targeted synthetic data for it (\S\ref{sec:bridge}), add it to the pool, and keep it only if the targeted benchmarks improve without regressing the others.

\paragraph{Resulting mixture.}
Table~\ref{tab:pretraining-data-sources} and Table~\ref{table:mid-train-data} give the full composition and mixing ratios. Generic web data (Dolma~3 CC, FineWeb-Edu, and DCLM) makes up $59.2\%$ of the mixture; the remainder comes from capability-targeted and synthetic sources.

\begin{table}[t]
\small
\centering
\caption{\small Pre-training datasets and mixture ratios. Dataset names link to their public sources.}
\vspace{-0.5em}
\label{tab:pretraining-data-sources}
\begingroup
\setlength{\tabcolsep}{1.5mm}
\renewcommand{\arraystretch}{1.02}
\begin{tabular}{@{}llr@{}}
\toprule
\textbf{Dataset} & \textbf{Subset} & \textbf{Mix} \\
\midrule
\href{https://huggingface.co/datasets/HuggingFaceFW/fineweb-edu}{FineWeb-Edu} & v1.4.0 & 0.86\% \\
\href{https://huggingface.co/datasets/allenai/dolma3_mix-6T-1025-7B}{Dolma 3} & Common Crawl & 57.84\% \\
 & Stack Edu & 2.95\% \\
 & olmOCR Science & 10.28\% \\
 & ProofPile Arxiv & 0.31\% \\
 & Wiki & 0.79\% \\
\href{https://huggingface.co/datasets/allenai/dolma}{Dolma} & FLAN & 0.58\% \\
\href{https://huggingface.co/datasets/mlfoundations/dclm-baseline-1.0}{DCLM} & Baseline & 0.51\% \\
\href{https://huggingface.co/datasets/IFM/MegaMath}{MegaMath} & \href{https://huggingface.co/datasets/IFM/MegaMath/tree/main/megamath-text-code-block}{Text and code blocks} & 0.69\% \\
 & \href{https://huggingface.co/datasets/IFM/MegaMath/tree/main/megamath-translated-code}{Translated code} & 0.27\% \\
 & \href{https://huggingface.co/datasets/IFM/MegaMath/tree/main/megamath-qa}{QA} & 0.98\% \\
 & \href{https://huggingface.co/datasets/IFM/MegaMath/tree/main/megamath-web-pro}{Web Pro} & 0.73\% \\
\href{https://huggingface.co/datasets/nvidia/Llama-Nemotron-Post-Training-Dataset}{Nemotron} & \href{https://huggingface.co/datasets/nvidia/Llama-Nemotron-Post-Training-Dataset/blob/main/SFT/code/code_v1.1.jsonl}{Code} & 0.51\% \\
 & \href{https://huggingface.co/datasets/nvidia/Llama-Nemotron-Post-Training-Dataset/blob/main/SFT/math/math_v1.1.jsonl}{Math} & 0.95\% \\
 & \href{https://huggingface.co/datasets/nvidia/Llama-Nemotron-Post-Training-Dataset/blob/main/SFT/science/science.jsonl}{Science} & 0.55\% \\
\href{https://huggingface.co/datasets/HuggingFaceTB/finemath}{FineMath} & Base and subsets & 0.33\% \\
\href{https://huggingface.co/datasets/open-web-math/open-web-math}{OpenWebMath} & Proof-Pile 2 & 0.34\% \\
\href{https://huggingface.co/datasets/EleutherAI/proof-pile-2/tree/main/algebraic-stack}{Algebraic Stack} & Proof-Pile 2 & 0.31\% \\
\href{https://huggingface.co/datasets/togethercomputer/RedPajama-Data-1T/blob/main/urls/arxiv.txt}{Arxiv} & Arxiv & 0.80\% \\
\href{https://data.together.xyz/redpajama-data-1T/v1.0.0/stackexchange/stackexchange.jsonl}{StackExchange} & StackExchange & 0.29\% \\
\href{https://huggingface.co/datasets/allenai/dolmino-mix-1124/tree/main/data/wiki}{Wiki} & English & 0.46\% \\
\href{https://huggingface.co/datasets/facebook/natural_reasoning}{Natural Reasoning} & -- & 0.73\% \\
\href{https://huggingface.co/datasets/HuggingFaceTB/cosmopedia}{Cosmopedia} & SmolLM & 0.27\% \\
\midrule
Bridge data (\S\ref{sec:bridge}) & Synthetic cross-domain & 15.78\% \\
Others & -- & 1.89\% \\
\midrule
\textbf{Total} & & \textbf{100.00\%} \\
\bottomrule
\end{tabular}
\endgroup
\end{table}

\begin{table}[t!]
\small
\centering
\caption{\small Mid-training datasets and mixture ratios, with 105B tokens sampled per phase according to the mix ratio.}
\vspace{-0.5em}
\label{table:mid-train-data}
\setlength{\tabcolsep}{1.5mm}
\renewcommand{\arraystretch}{1.02}
\begin{tabular}{l l r r r}
\toprule
\textbf{Dataset} & \textbf{Subset} & \textbf{Rows (M)} & \textbf{Phase1 Mix} & \textbf{Phase2 Mix} \\
\midrule
\multirow{6}{*}{\href{https://huggingface.co/datasets/allenai/dolmino-mix-1124}{Dolmino}}
& DCLM Baseline & 606 & 37.03\% & 6.58\% \\
& FLAN & 57.3 & 4.10\% & 0.73\% \\
& peS2o & 38.8 & 11.41\% & 2.03\% \\
& Wiki & 6.17 & 2.66\% & 0.47\% \\
& StackExchange & 2.48 & 2.12\% & 2.02\% \\
& Math & 21 & 11.63\% & 29.39\% \\
\midrule
\multirow{2}{*}{\href{https://huggingface.co/datasets/nvidia/Llama-Nemotron-Post-Training-Dataset}{Nemotron}}
& \href{https://huggingface.co/datasets/nvidia/Llama-Nemotron-Post-Training-Dataset/tree/main/SFT/code}{Code (SFT)} & 10.1 & 5.69\% & 8.08\% \\
& \href{https://huggingface.co/datasets/nvidia/Llama-Nemotron-Post-Training-Dataset/tree/main/SFT/math}{Math (SFT)} & 22.1 & 3.45\% & 19.59\% \\
\midrule
\multirow{1}{*}{StarCoder}
& \href{https://huggingface.co/datasets/bigcode/starcoderdata}{StarCoder} & 206 & 6.90\% & 9.80\% \\
\midrule
\multirow{1}{*}{Dolma 3}
& \href{https://huggingface.co/datasets/allenai/dolma3_mix-6T-1025-7B}{Stack-Edu (Python, C++, C, JavaScript)} & 526 & 15.00\% & 21.31\% \\
\midrule
\textbf{Total} &  &  & \textbf{100.00\%} & \textbf{100.00\%} \\
\bottomrule
\end{tabular}
\end{table}

\subsection{Synthetic Cross-domain Bridge Data}
\label{sec:bridge}

Data mixture optimization must account for capability domains that do not combine additively. Adding data for one capability can affect performance on others \citep{xie2023doremi,aryabumi2025code,dong2024abilities}, creating trade-offs as training data composition and scale vary. These cross-domain trade-offs are particularly pronounced in small language models, where limited capacity constrains the ability to accommodate diverse capabilities. This challenge is especially relevant to our setting, which requires a compact model to support both multi-turn tool use and long-horizon single-turn reasoning. Simply increasing the mixture weight of an underperforming domain may yield diminishing gains while degrading other capabilities. To mitigate these trade-offs, we introduce synthetic cross-domain \emph{bridge} data directly into pre-training.

Bridge documents connect otherwise competing domains by combining their content in a natural intermediate form. We use three bridge types: natural-language explanations of code, mathematical problems presented as factual or scientific text, and reasoning over encyclopedic content. We generate {92.4M} bridge documents ({108}B unique tokens) with Llama-3 8B~\citep{grattafiori2024llama} from seed documents drawn from mathematical (FineMath, OpenWebMath, MegaMath, Algebraic Stack), question-answering (StackExchange), encyclopedic (Wikipedia), and educational (SmolLM corpus) sources, using one fixed prompt template per bridge type.
We sample at temperature {0.8} and filter outputs by {length (200 to 4{,}000 tokens), $n$-gram repetition, and a model-judged fidelity check against the seed document}. The generated documents are then deduplicated and decontaminated like every other source (\S\ref{sec:data-processing}). Bridge data receives $15.8\%$ of the pre-training mixture, about 189B of the 1.2T pre-training tokens.

Bridge data does not introduce a new capability. Instead, it narrows the distributional gap between existing ones, which helps the model acquire them jointly under a limited token budget. In our ablation, bridge data improves all 19 evaluated benchmarks, yielding an average absolute gain of 1.27 percentage points (3.48\% relative), as detailed in Table~\ref{tab:bridge_ablation}.

\subsection{Monitoring the Full-scale Run with a Readiness Predictor}
\label{sec:readiness}

The resulting best mixture, $w^\star$, is used for the full-scale pre-training run, which is monitored by the readiness predictor.
The selection criterion combines two signals: zero-shot accuracy averaged across capability categories, and the sampling lift on generation tasks.

\begin{figure}[t]
\centering
\includegraphics[width=\textwidth]{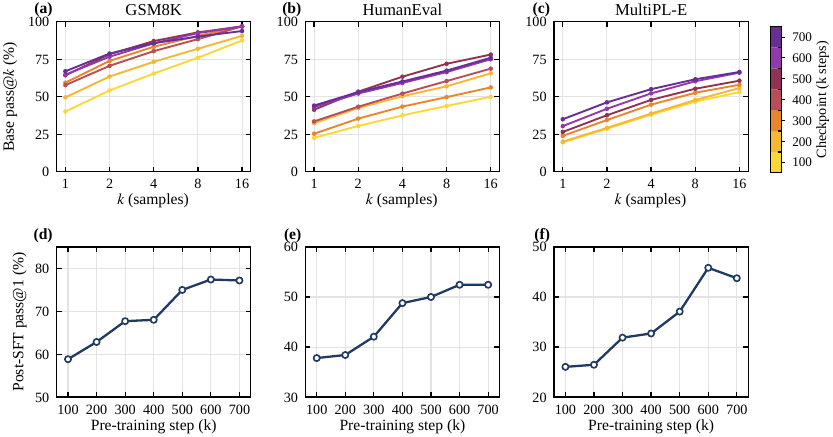}
\caption{\small \textit{Sampling lift} provides an early signal of post-training potential.
Later in pre-training, $\operatorname{pass}@1$ keeps improving while $\operatorname{pass}@16$ saturates or declines, shrinking the sampling lift $\Delta_{16}=\operatorname{pass}@16-\operatorname{pass}@1$.
This indicates that less solvable capability remains beyond the model's most likely output.
This shrinkage coincides with diminishing post-training gains: checkpoints with a plateauing $\Delta_{16}$ improve little after post-training. The data-improvement loop therefore uses $\Delta_{16}$ as a predictor input.}
\label{fig:sampling-lift}
\end{figure}

\subsubsection{Sampling lift on generation tasks}
Single-sample accuracy measures how reliably a model produces a correct solution.
It does not capture whether correct solutions remain reachable under repeated sampling.
To separate the two, we track the \emph{sampling lift} at checkpoint $t$:
\begin{equation}
\Delta_{k,t} \;=\; \operatorname{pass}@k_t - \operatorname{pass}@1_t ,
\label{eq:sampling-lift}
\end{equation}
which is the additional success obtained from $k$ samples rather than one \citep{chen2021evaluating}.
We use $k{=}16$ throughout.
As shown in Figure~\ref{fig:sampling-lift}, early checkpoints improve rapidly on both $\operatorname{pass}@1$ and $\operatorname{pass}@16$.
Sampling lift is not a direct measure of behavioral diversity.
It measures how much solvable capability lies outside the model's most likely output, which is the capability post-training can draw on.

\subsubsection{Category-level averages for mixture ranking}
Let $x_{t,b}$ denote the zero-shot score of checkpoint $t$ on benchmark $b$, for the 20 benchmarks of Table~\ref{tab:readiness-benchmarks} grouped into the set $\mathcal{C}$ of six capability categories.
We first average within each category and then across categories:
\begin{equation}
\bar{x}_t \;=\; \frac{1}{|\mathcal{C}|} \sum_{c \in \mathcal{C}} \frac{1}{|\mathcal{E}_c|} \sum_{b \in \mathcal{E}_c} x_{t,b},
\label{eq:macro-avg}
\end{equation}
where $\mathcal{E}_c$ is the set of benchmarks in category $c$.
Every category therefore carries equal weight, regardless of how many benchmarks it contains, so the four knowledge benchmarks count no more than the two mathematics benchmarks.
A candidate mixture is preferred when it raises $\bar{x}_t$ without reducing sampling lift $\Delta_{16,t}$ at matched training steps.

\begin{figure}[t]
  \centering
  \includegraphics[width=\linewidth]{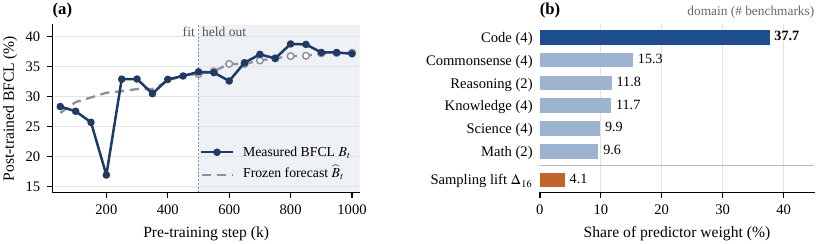}
  \caption{\small \textbf{Zero-shot capability composition predicts post-training performance.}
  (a)~Post-trained BFCL measured at pre-training checkpoints, compared with the predictor's forecast $\widehat{B}_t$. The predictor is fit only on checkpoints before 500k steps, using their zero-shot benchmark scores and sampling lift, then frozen and applied to the later, held-out checkpoints (shaded).
  (b)~Fitted predictor weights $\psi_c$ by capability domain and for sampling lift $\Delta_{16}$, computed as normalized absolute ridge coefficients. Larger shares indicate a stronger predictive association, not a causal effect. Parentheses give the number of benchmarks per domain.}
  \label{fig:readiness-prediction}
  \label{fig:readiness-domain-contributions}
\end{figure}

The category average is used only to rank mixtures against each other at the same training step, where it is a fair comparison. It is not used to judge how far a single run has progressed: as Figure~\ref{fig:readiness-prediction} shows, the average can keep rising along a run while post-training performance stalls.
That second question is answered by a separate readiness predictor.

\subsubsection{A readiness predictor from zero-shot scores}
Agentic capabilities such as multi-turn planning, tool use, and error recovery are typically realized through post-training. Their effectiveness, however, depends on foundations laid during pre-training, including reasoning, coding, and problem solving. This creates a fundamental difficulty: \textit{the capabilities that determine a model's eventual agentic performance may already be developing during pre-training, yet they cannot be reliably assessed through direct agentic evaluation at that stage}. The difficulty is sharpest for a long pre-training run, where the practical question is when further pre-training stops paying off after post-training. We therefore ask: \textit{can signals observed during pre-training predict a model's agentic performance after post-training?}

Let $\mathbf{x}_t\in\mathbb{R}^{20}$ collect the 20 zero-shot scores of checkpoint $t$ (Table~\ref{tab:readiness-benchmarks}).
We sample checkpoints every 50k steps, apply the same fixed post-training recipe to each, and measure BFCL, a direct measure of downstream tool-use capability.
The resulting pairs form a calibration set $\mathcal{K}=\{(\mathbf{x}_t,\Delta_{16,t},B_t)\}$, where $B_t$ is the post-trained BFCL score, and we fit the \emph{readiness predictor} once:
\begin{equation}
\widehat{B}_t \;=\; R(\mathbf{x}_t,\Delta_{16,t}),
\label{eq:posttrain-prediction}
\end{equation}
where $R$ is a ridge regression.
We fit it only on checkpoints before 500k steps and evaluate it on later checkpoints whose post-training scores are never used in fitting.
Figure~\ref{fig:readiness-prediction} shows that, on the held-out checkpoints of the baseline mixture, the prediction tracks measured post-training performance.
Unlike the category average of Eq.~\ref{eq:macro-avg}, $R$ learns its own weight for each benchmark, so it can register when the capabilities that matter for post-training stop improving even if the average keeps rising.
Once fit, $R$ costs one zero-shot evaluation pass plus $\operatorname{pass}@16$ sampling per checkpoint, with no post-training.
We apply it to the checkpoints of the full-scale run and treat a plateau in $\widehat{B}_t$ as the sign that the run has reached diminishing returns.

\subsubsection{Which pre-trained capabilities matter?}
We compute the normalized contribution $\psi_c$ of each capability category $c$ (and of $\Delta_{16}$) to $R$ through a coefficient-based analysis (Figure~\ref{fig:readiness-domain-contributions}).
These contributions should be read as predictive associations, not causal effects.
Two findings emerge.
First, a benchmark's predictive value depends on how closely it aligns with the downstream objective.
Code provides the strongest signal ($\psi_{\text{code}}=37.7\%$), plausibly because its compositional, constraint-driven, and executable structure resembles tool-use behavior.
Other capabilities contribute complementary but more indirect signal, and highly correlated benchmarks add little beyond one another.
Second, sampling lift carries information that zero-shot accuracy does not, since it reflects the solvable capability available beyond the model's most likely output.
As training progresses, the gap indicated by the sampling lift narrows, particularly as $\operatorname{pass}@16$ approaches saturation, and downstream SFT performance shows diminishing returns over the same period.
This is consistent with evidence that RL post-training largely re-weights solutions the base model can already sample rather than creating new ones \citep{yue2025doesrl}.
These contributions motivate our mid-training strategy, which concentrates additional tokens on code, mathematics, instruction following, and tool use.
They also show that the equal category weighting used for mixture selection differs from the weighting that best predicts post-training performance; weighting the selection criterion by $\psi$ is a natural extension that we leave to future work.
Together, these components make pre-training and mid-training produce more than a stronger base model: they produce a \emph{post-training-ready substrate} whose capabilities are broad, balanced, and aligned with the behaviors post-training can most effectively unlock.

\section{Eliciting Reasoning and Agentic Capabilities through Post-Training}
\label{sec:post_training}

Our post-training pipeline consists of three stages: (1) general supervised fine-tuning (SFT), (2) agentic reasoning SFT, and (3) on-policy knowledge distillation (OPD). Beyond the individual training stages, we investigate two critical factors for effective post-training: the joint training of reasoning and agentic capabilities and the alignment of chat templates and trajectory distributions across SFT and OPD. Our analysis reveals important interactions between data formatting, behavioral distributions, and training objectives, offering practical insights into building small language models with both strong reasoning and agentic capabilities.

\subsection{Supervised Fine-Tuning}
\label{sec:post_sft}

Post-training turns MobileLLM-R2-Base model into MobileLLM-R2, a model that reasons over long horizons and
acts as an agent with tools. Our SFT pipeline consists of two stages. Stage 1 performs general-purpose instruction tuning, while Stage 2 jointly trains on long-horizon reasoning and multi-turn agentic trajectories to develop complementary capabilities within a single model. Beyond data composition, we also adapt Stage 2 SFT to improve compatibility with subsequent on-policy distillation (OPD). We first examine the effects of jointly training reasoning and agentic capabilities, and revisit the OPD-oriented adaptations after introducing the distillation framework.

\subsubsection{Unifying long-horizon reasoning and agentic capabilities}
\label{sec:effort_routing}

Long-horizon reasoning and interactive agentic tasks impose distinct computational demands. Mathematical and coding problems often benefit from extended deliberation before producing a solution, whereas agentic tasks require interleaving reasoning with environment interactions, emphasizing contextual retrieval, state tracking, and action selection. Integrating these complementary capabilities within a compact language model presents an important challenge: can heterogeneous reasoning objectives be jointly optimized without compromising performance on either task family?

We investigate this question by jointly training on long-horizon reasoning and multi-turn agentic trajectories. As shown in Table~\ref{tab:low_high_training_data}, training exclusively on long-horizon reasoning data achieves strong mathematical and coding performance but fails on SWE-Bench Verified (0.0\%), while agentic-only training achieves 25.4\% on SWE-Bench Verified at the expense of substantial degradation in mathematical and coding benchmarks. In contrast, joint training achieves 91.2\% on MATH500, 37.1\% on AIME'25, and 29.3\% on LCB, maintaining performance comparable to that of the model trained exclusively on long-horizon mathematical and coding reasoning data (90.8\%, 38.3\%, and 31.4\%, respectively), while further improving SWE-Bench Verified accuracy to 27.0\%. These results demonstrate that long-horizon reasoning and agentic supervision can be effectively integrated, substantially broadening the model's capabilities with minimal interference between the two objectives.

Beyond benchmark accuracy, we examine how the jointly trained model adapts its reasoning behavior to different task demands. Table~\ref{tab:effort_analysis} reveals substantial variation in generation lengths across tasks: median output lengths range from 1,724 tokens on MATH500 to 13,157 on AIME'25 and 16,718 on LCB, compared with only 210 tokens per interaction turn on SWE-Bench Verified. This contrast reflects the distinct structures of the underlying tasks: extended single-turn deliberation for complex reasoning problems versus shorter reasoning steps interleaved with tool interactions for agentic workflows. Together, these findings demonstrate that joint post-training enables a compact model with 2.5B parameters to accommodate heterogeneous reasoning patterns while preserving strong performance across both long-horizon reasoning and interactive agentic tasks.

\begin{table}[t]
\small
\centering
\caption{\small Performance comparison under different training data mixtures. }
\vspace{-0.5em}
\label{tab:low_high_training_data}
\setlength{\tabcolsep}{1.5mm}
\renewcommand{\arraystretch}{1.02}
\begin{tabular}{@{}lccccc@{}}
\toprule
\textbf{Training data} & \textbf{MATH500} & \textbf{AIME'25} & \textbf{LCB} & \textbf{SWE-Bench Verified} \\
\midrule
Long-horizon reasoning data only & 90.8\% & 38.3\% & 31.4\% & 0.0\% \\
Agentic multi-turn data only & 30.0\% & 2.6\% & 12.0\% & 25.4\% \\
Joint reasoning + agentic training & 91.2\% & 37.1\% & 29.3\% & 27.0\% \\
\bottomrule
\end{tabular}
\end{table}

\begin{table}[t]
\small
\centering
\caption{\small Output length distributions across long-horizon reasoning and multi-turn agentic tasks. For SWE-Bench Verified, output length is measured per interaction turn.}
\vspace{-0.5em}
\label{tab:effort_analysis}
\setlength{\tabcolsep}{1.5mm}
\renewcommand{\arraystretch}{1.02}
\begin{tabular}{@{}lccc|c@{}}
\toprule
& \multicolumn{3}{c|}{\textbf{Long-horizon Single-turn Reasoning}} & \textbf{Agentic Multi-turn Tool Use} \\
\noalign{\vspace{0.2em}}
\textbf{Metric} & MATH500 & AIME'25 & LCB &  SWE-Bench Verified \\
\midrule
Output length (p50) & 1724 & 13157 & 16718 & 210 \\
\midrule
Output length (p90) & 8331 & 21527 & 29999 & 629 \\
\bottomrule
\end{tabular}
\end{table}

\subsection{On-Policy Distillation}
\label{sec:post_opd}

Naive on-policy knowledge distillation (OPD) trains the student on sequences it generates itself, with the teacher's next-token distribution as the target at every position. For each prompt $x$ we sample one rollout $y$ from the student $\pi_\theta$ and minimize the forward KL divergence from the teacher $\pi_T$ to the student at each position:
\begin{equation}
  \mathcal{L}_{\mathrm{OPD}}
  = \mathbb{E}_{x,\; y\sim\pi_\theta(\cdot\mid x)}
    \sum_{t=1}^{|y|} \mathrm{KL}\bigl(\pi_T(\cdot\mid x, y_{<t}) \,\big\|\, \pi_\theta(\cdot\mid x, y_{<t})\bigr).
\end{equation}

Because OPD supervises student-generated trajectories, its effectiveness depends not only on the distillation objective but also on the compatibility of these trajectories with the teacher. We examine this compatibility along two dimensions: the structural alignment of chat templates and the distributional alignment of model behaviors.

\subsubsection{Two dimensions of alignment: chat templates and trajectory distributions}
\label{sec:chat_template_contract}

\textbf{Chat Templates.} Tool-use post-training requires more than learning to map user requests to structured calls. It also requires a consistent protocol for serializing interaction trajectories into tokens and parsing generated tokens into executable actions. We find that seemingly minor protocol mismatches can substantially degrade tool-use performance or invalidate entire interaction episodes.

Let $M$ denote a structured interaction trajectory containing roles, channels, recipients, reasoning traces, tool calls, and observations. A chat template $\tau$ defines a renderer $R_\tau$ that serializes the trajectory into tokens, a supervision mask function $m_\tau$ that specifies which tokens contribute to the training objective, and an inference parser $P_\tau$ that interprets generated tokens as structured actions:
\begin{equation}
\begin{aligned}
(x,m) &= (R_\tau(M),m_\tau(M)),\\
\mathcal{L}_{\mathrm{SFT}}(\theta) &= -\sum_{t=1}^{T}m_t\log p_\theta(x_t\mid x_{<t}),\\
\hat{a} &= P_\tau(\hat{x}), \quad
\hat{x}\sim p_\theta(\cdot\mid x_{\mathrm{ctx}}).
\end{aligned}
\end{equation}
Here, $x=(x_1,\ldots,x_T)$ is the rendered token sequence, $m\in\{0,1\}^T$ is the corresponding token-level supervision mask, and $p_\theta$ denotes the model's autoregressive distribution. At inference time, $x_{\mathrm{ctx}}$ denotes the rendered interaction history, $\hat{x}$ the generated token sequence, and $\hat{a}$ the resulting structured action.

A chat template therefore defines not only the representation of interaction trajectories but also the supervision and execution semantics of the model. Effective training requires consistency across these three components: message boundaries and control tokens must preserve their intended semantics, loss masks must supervise the correct assistant decisions, and inference-time parsing and stopping rules must match the training protocol. Violations can silently alter the learning objective even when the serialized data appears syntactically valid.

\textbf{Protocol incompatibility.}
\begin{figure*}[t]
\centering

\begin{minipage}[t]{0.32\textwidth}
\centering
\textbf{LLaMA 3.1}\par\vspace{4pt}
\begin{protocolbox}
<|start_header_id|>assistant<|end_header_id|>

<|python_tag|>\{
  "type": "function",
  "name": "get_weather",
  "parameters": \{"city": "Tokyo"\}
\}<|eom_id|>

<|start_header_id|>ipython<|end_header_id|>

...<|eot_id|>

<|start_header_id|>assistant<|end_header_id|>

...<|eot_id|>
\end{protocolbox}
\end{minipage}
\hfill
\begin{minipage}[t]{0.32\textwidth}
\centering
\textbf{Qwen3.5}\par\vspace{4pt}
\begin{protocolbox}
<|im_start|>assistant
<tool_call>
<function=get_weather>
<parameter=city>
Tokyo
</parameter>
</function>
</tool_call><|im_end|>

<|im_start|>user
<tool_response>
...
</tool_response><|im_end|>

<|im_start|>assistant
...<|im_end|>
\end{protocolbox}
\end{minipage}
\hfill
\begin{minipage}[t]{0.32\textwidth}
\centering
\textbf{Muse-Glimmer}\par\vspace{4pt}
\begin{protocolbox}
<|start|>assistant to=self<|message|>
...<|eom|>

<|start|>assistant to=get_weather<|message|>
<atem:function_calls>
<atem:invoke name="get_weather">
<atem:parameter name="city">Tokyo</atem:parameter>
</atem:invoke>
</atem:function_calls><|eot|>

<|start|>tool<|message|>
<tool_output>...</tool_output><|eot|>

<|start|>assistant to=user<|message|>
...<|eot|>
\end{protocolbox}
\end{minipage}

\caption{Comparison of tool-use serialization across LLaMA 3.1, Qwen3.5, and Muse-Glimmer. The protocols differ in message boundaries, tool-call representation, routing semantics, and termination tokens, requiring explicit adaptation when transferring interaction trajectories across model families.}
\label{fig:chat_template_comparison}
\end{figure*}

Although different chat templates may encode semantically equivalent interactions, their serialized representations are not interchangeable due to differences in control tokens, routing semantics, and termination rules. Cross-model transfer therefore requires adapting structured trajectories to the target model's native protocol, including consistent supervision masks and inference-time parsing.
As illustrated in Figure~\ref{fig:chat_template_comparison}, Muse-Glimmer adopts an explicit recipient-based routing protocol that distinguishes internal reasoning (\cttok{to=self}), tool invocation (\cttok{to=get_weather}), and user-facing responses (\cttok{to=user}). Unlike LLaMA 3.1 and Qwen3.5, which encode tool interactions through different message structures, Muse-Glimmer makes these transitions explicit through routing headers and distinct message termination tokens. Since many existing agentic datasets were constructed using earlier chat templates, directly reusing their serialized trajectories can introduce mismatches in routing, message boundaries, and stopping behavior. We therefore convert source trajectories into the native Muse-Glimmer protocol, preserving role and recipient semantics, tool-call structure, and termination rules.

\label{sec:template_opd_routing}

\textbf{Trajectory distribution alignment.} Beyond structural compatibility of chat templates, we find that trajectory distribution alignment between SFT data and the OPD teacher is equally critical. SFT trajectories encode specific behavioral patterns, which may differ substantially from how the OPD teacher solves the same task. These patterns persist in the rollouts of students trained on such data, inducing a systematic teacher–student distribution mismatch that limits the effectiveness of subsequent on-policy distillation.

To mitigate this distribution mismatch, we incorporate teacher-rewritten solutions into the SFT data. By exposing the student to the behavioral patterns of the OPD teacher during SFT, we encourage the student to generate trajectories that are better aligned with the teacher's solution distribution, providing a more favorable initialization for subsequent on-policy distillation.

While rewriting solutions from the initial prompt improves distributional alignment, multi-turn agentic tasks introduce an additional challenge: each student action changes the interaction state and consequently the context under which subsequent decisions are made. Teacher-generated trajectories starting from the initial prompt do not explicitly provide supervision for the intermediate states induced by the student's own actions. Instead, we find that Till-End strategy, in which the teacher takes over from intermediate states sampled from student-generated trajectories and continues the interaction until termination, proves particularly effective for multi-turn agentic tasks. Detailed ablation of different teacher intervention strategies for agentic behavior can be found in Appendix~\ref{sec:opd_agentic}.

Importantly, although Till-End strategy operates on states sampled from the student policy, its supervision takes the form of teacher-generated continuations optimized through cross-entropy rather than per-token KL divergence. It can therefore be viewed as a generalization of teacher rewriting from initial prompts to student-induced intermediate states. Since our training pipeline jointly develops single-turn reasoning and multi-turn agentic capabilities, we seek to maintain a consistent optimization objective across task domains within each training stage to promote training stability. We therefore incorporate the teacher-completed trajectories from Till-End strategy into Stage 2 SFT, alongside conventional teacher-rewritten solutions, under a unified cross-entropy objective, and reserve Stage 3 for KL-based on-policy distillation. The complete post-training recipe, including data composition and teacher-written solution proportions, is summarized in Table~\ref{tab:post-training_data}.

\begin{table}[t]
\small
\centering
\caption{\small Impact of aligning SFT data with the target chat template and incorporating teacher-written solutions on subsequent OPD. Results report mean accuracy across AIME'24 and AIME'25 before and after distillation.}
\label{tab:opd_start}
\setlength{\tabcolsep}{1.5mm}
\renewcommand{\arraystretch}{1.02}
\vspace{-0.5em}
\begingroup
\begin{tabular}{@{}lcc@{}}
\toprule
\textbf{SFT Stage 2 data format} & \textbf{After SFT} & \textbf{After OPD} \\
\midrule
Original chat template & 43.4 & 17.1 \\
Muse-Glimmer template & \textbf{49.0} & 36.8 \\
+ 25\% teacher-written solutions & 43.9 & \textbf{54.9} \\
\bottomrule
\end{tabular}
\endgroup
\end{table}

\begin{table}[t]
\footnotesize
\centering
\caption{\small Post-training datasets and training configurations. Objective specifies the training loss used at each stage: cross-entropy (CE) for SFT and per-token Kullback--Leibler (KL) divergence for OPD. Rows indicate the number of examples before dataset-level upsampling, while stage totals reflect the effective mixture size after applying repeat factors. For OPD, they represent retained on-policy trajectories generated from prompts sampled from the corresponding source dataset. Repeat factors specify dataset-level upsampling before mixture construction, while epochs indicate training passes over the resulting mixture. Teacher-rewritten denotes the fraction of examples whose solutions are replaced with teacher-generated targets. $^{*}$For OpenThoughts-Agent-SFT, teacher rewriting follows the Till-End strategy, in which the teacher completes a trajectory from a student-visited state.}
\label{tab:post-training_data}
\vspace{-0.5em}
\begingroup
\setlength{\tabcolsep}{1.5mm}
\renewcommand{\arraystretch}{1.02}
\begin{tabular}{@{}llllrrrr@{}}
\toprule
\textbf{Training Stage} & \textbf{Objective} & \textbf{Dataset} & \textbf{Data Type} & \textbf{Rows} & \textbf{Repeat} & \textbf{Epochs} & \textbf{Teacher-rewritten} \\
\midrule
\multirow{3}{*}{\shortstack[l]{Stage 1:\\General SFT}}
  & \multirow{3}{*}{CE}
  & \href{https://huggingface.co/datasets/allenai/Dolci-Think-SFT-7B}{Dolci-Think-SFT-7B} & General Reasoning & 2.3M & 1$\times$ & 2 & 0\% \\
  & & \href{https://huggingface.co/datasets/nvidia/Nemotron-Agentic-v1}{Nemotron-Agentic-v1} & Agentic & 100K & 1$\times$ & 2 & 0\% \\
  & & \textbf{Total} & & \multicolumn{2}{c}{\textbf{2.4M}} & -- & --  \\
\midrule
\multirow{6}{*}{\shortstack[l]{Stage 2:\\Agentic \\ Reasoning \\ SFT}}
  & \multirow{6}{*}{CE}
  & \href{https://huggingface.co/datasets/nvidia/OpenMathReasoning}{OpenMathReasoning (CoT)} & Math Reasoning & 3.2M & 1$\times$ & 1 & 25.0\% \\
  & & \href{https://huggingface.co/datasets/nvidia/OpenCodeReasoning-2}{OpenCodeReasoning-2} & Code Reasoning & 2.6M & 1$\times$ & 1 & 12.9\% \\
  & & \href{https://huggingface.co/datasets/nvidia/OpenScienceReasoning-2}{OpenScienceReasoning-2} & Science Reasoning & 1.6M & 1$\times$ & 1 & 24.8\% \\
  & & \href{https://huggingface.co/datasets/open-thoughts/OpenThoughts-Agent-SFT-100K}{OpenThoughts-Agent-SFT} & Agentic & 94K & 14$\times$ & 1 & 5.0\%$^{*}$ \\
  & & Synthesized Agentic Data & Agentic & 27K & 3$\times$ & 1 & 0\% \\
  & & \textbf{Total} & & \multicolumn{2}{c}{\textbf{8.8M}} & -- & -- \\
\midrule
\multirow{4}{*}{\shortstack[l]{Stage 3:\\OPD}}
  & \multirow{4}{*}{KL}
  & \href{https://huggingface.co/datasets/nvidia/OpenMathReasoning}{OpenMathReasoning} & Math Reasoning & 561K & 1$\times$ & 1 & -- \\
  & & \href{https://huggingface.co/datasets/nvidia/OpenCodeReasoning-2}{OpenCodeReasoning-2} & Code Reasoning & 184K & 1$\times$ & 1 & -- \\
  & & \href{https://huggingface.co/datasets/open-thoughts/OpenThoughts-Agent-SFT-100K}{OpenThoughts-Agent} & Agentic & 66K & 1$\times$ & 1 & -- \\
  & & \textbf{Total} & & \multicolumn{2}{c}{\textbf{812K}} & -- & -- \\
\bottomrule
\end{tabular}
\endgroup
\end{table}

\textbf{Results.} Table~\ref{tab:opd_start} demonstrates that mismatches with limited impact on SFT accuracy can have substantially larger consequences during OPD. Adapting the SFT data to the target chat template substantially mitigates the degradation after OPD (17.1 to 36.8). Nevertheless, template alignment alone is insufficient: incorporating 25\% teacher-written solutions enables OPD to improve performance to 54.9.

These results highlight the importance of both chat-template alignment and trajectory-distribution alignment for effective OPD. While their effects are less pronounced during SFT, both become substantially more consequential during OPD, where the teacher supervises student-generated trajectories. This suggests that effective OPD requires not only structurally compatible interaction protocols but also student trajectories that are well aligned with the teacher's behavioral distribution.

\begin{table}[!t]
\small
\centering
\caption{\small Effect of iterative OPD on mathematical reasoning. Mean accuracy on AIME'24 and AIME'25 across successive rounds of math OPD. Teacher-rewritten SFT enables strong gains in a single round, whereas standard SFT requires multiple rounds to reach comparable performance. }
\label{tab:abl_rounds}
\vspace{-0.5em}
\begingroup
\setlength{\tabcolsep}{1.5mm}
\renewcommand{\arraystretch}{1.02}
\begin{tabular}{@{}lcccccc@{}}
\toprule
\textbf{Starting point} & \textbf{Before OPD} & \textbf{Round 1} & \textbf{Round 2} & \textbf{Round 3} & \textbf{Round 4} & \textbf{Round 5} \\
\midrule
SFT with teacher-rewritten solutions & 43.9 & 54.9 & 53.3 & 53.9 & 53.8 & 53.6 \\
SFT without teacher-written solutions & 41.2 & 26.7 & 43.8 & 48.8 & 54.2 & 53.8 \\
\bottomrule
\end{tabular}
\endgroup
\end{table}

\textbf{Iterative OPD.} We further validate the importance of teacher-written solutions through iterative OPD. As shown in Table~\ref{tab:abl_rounds}, without teacher-written solutions, OPD initially degrades performance and requires multiple rounds to recover. In contrast, incorporating them enables a single round to improve mean AIME accuracy from 43.9\% to 54.9\%, with no further gains from additional rounds. We therefore adopt a single round of OPD, using rollouts generated by the Stage 2 SFT model and training on the resulting trajectories for one epoch.

\section{Main Results}
\label{sec:main_results}
\begin{table*}[t]
\scriptsize
\centering
\caption{\small{Performance comparison of \ours{} with six state-of-the-art open-source language models ranging from sub-billion to 7B parameters: MobileLLM-R1.5~\citep{zhao2026mobilellm}, Qwen3.5-2B~\citep{team2026qwen3}, SmolLM3-3B~\citep{bakouch2025smollm3}, Nemotron-3-Nano-4B~\citep{blakeman2025nvidia}, Gemma-4-E2B~\citep{team2026gemma}, and Olmo-3-7B-Instruct~\citep{olmo2025olmo}. All results are reproduced from publicly released checkpoints under a unified evaluation framework and protocol, and generally align with official reports. Minor discrepancies, particularly for OLMo-3, may arise from differences between its official evaluation framework (\href{https://github.com/allenai/olmes/tree/main}{OLMES}) and ours. We report benchmark accuracy in percent (\%). Among models with fewer than 3.5B total parameters, bold denotes results within 1.0 percentage point of the best result in this size group. Underlining denotes the best result across all models.}}
\label{tab:overall_acc}
\vspace{-0.5em}
\setlength{\tabcolsep}{0.9mm}
\renewcommand{\arraystretch}{1.02}
\begin{tabular}{llcccccc|c}
\toprule
& & MobileLLM-R1.5- & Qwen3.5- & SmolLM3- & Nemotron-3- & Gemma-4- & Olmo-3- & $\ours{}$- \\
& & 950M & 2B & 3B & Nano-4B & E2B & 7B & 2.5B \\
\noalign{\vspace{0.2em}}
\hdashline
\noalign{\vspace{0.2em}}
Category & Benchmark & 0.95B & 2.27B & 3.08B & 3.97B & 2.00B / 5.12B & 7.30B & 2.50B \\
\midrule
Math
& AIME24 (avg@16) & 39.8 & 13.3 & 46.7 & 58.5 & 48.1 & 54.8 & \textbf{\underline{59.4}} \\
& AIME25 (avg@16) & 31.0 & 16.5 & 37.7 & \underline{56.5} & 36.0 & 42.3 & \textbf{49.0} \\
& MATH500 & 86.6 & 67.8 & \textbf{91.8} & 91.4 & 77.4 & \underline{92.0} & \textbf{91.0} \\
& GSM8K & 83.7 & 76.2 & 88.0 & 87.2 & 90.1 & \underline{93.3} & \textbf{91.9} \\
\midrule
Knowledge
& MMLU & 37.4 & \textbf{\underline{65.9}} & 63.2 & 65.8 & 58.1 & 63.4 & 56.5 \\
& GPQA & 29.3 & \textbf{51.5} & 41.4 & \underline{61.1} & 44.4 & 41.9 & 45.2 \\
\midrule
Coding
& HumanEval & 11.6 & 36.0 & 54.3 & \underline{72.0} & 67.1 & 47.6 & \textbf{69.5} \\
& MBPP & 5.6 & 23.2 & \textbf{\underline{50.0}} & 45.8 & 47.4 & 39.2 & \textbf{49.6} \\
& LiveCodeBench & 29.1 & 12.6 & 28.8 & 45.3 & 44.8 & 21.8 & \textbf{\underline{47.3}} \\
\midrule
Agentic
& BFCL V4 & 10.2 & 48.9 & 25.7 & \underline{53.8} & 41.9 & 31.8 & \textbf{50.6} \\
& SWE-Bench Verified & 0.0 & 15.6 & 0.0 & 7.3 & 1.1 & 0.6 & \textbf{\underline{30.9}} \\
\midrule
OOD
& Terminal-Bench & 0.0 & 4.1 & 0.4 & 2.6 & 1.9 & 1.1 & \textbf{\underline{5.2}} \\
& Aider Polyglot & 2.7 & 3.7 & 2.5 & 5.0 & 3.7 & 2.1 & \textbf{\underline{6.5}} \\
& BFCL-Parity & 0.0 & 30.1 & 70.2 & 68.0 & \underline{84.0} & 69.4 & \textbf{77.5} \\
\bottomrule
\end{tabular}
\end{table*}

Following the training pipeline in Figure~\ref{fig:overall_pipeline}, we obtain our final 2.5B-parameter model, \ours{}. Trained on 1.2T pre-training tokens and fewer than 2T tokens in total, \ours{} supports both long-horizon reasoning and multi-turn agentic tool use. This jointly designed training recipe offers a token-efficient approach to developing complementary reasoning and agentic capabilities within a compact model. We evaluate \ours{} against six state-of-the-art open-source models across mathematical reasoning, knowledge, coding, and agentic benchmarks, as shown in Table~\ref{tab:overall_acc}. Architecture and training details are provided in Section~\ref{sec:recipe}.

\textbf{Strong agentic capabilities at a compact scale.} The most notable result is on SWE-Bench Verified, where \ours{} achieves 30.9\%, outperforming the strongest baseline, Qwen3.5-2B (15.6\%), by 15.3 percentage points. This improvement is particularly notable given the demands of software engineering tasks, which require iterative code understanding, tool interaction, and multi-step problem solving. Beyond SWE-Bench Verified, \ours{} achieves competitive performance on BFCL V4 (50.6\%) and attains the highest scores among all evaluated models on Terminal-Bench (5.2\%) and Aider Polyglot (6.5\%). It also achieves 77.5\% on BFCL-Parity, outperforming all other models with fewer than 3.5B total parameters. These results suggest that the acquired agentic capabilities generalize beyond individual training and evaluation settings.

\textbf{Competitive reasoning and coding performance.} Importantly, strong agentic performance is accompanied by competitive mathematical reasoning and coding capabilities. \ours{} achieves the highest AIME24 accuracy (59.4\%) among all evaluated models, alongside strong results on AIME25 (49.0\%), MATH500 (91.0\%), and GSM8K (91.9\%). In coding, it reaches 69.5\% on HumanEval and achieves the highest LiveCodeBench accuracy (47.3\%), outperforming both Qwen3.5-2B and SmolLM3-3B on both benchmarks. Notably, \ours{} also surpasses the substantially larger OLMo-3-7B-Instruct on AIME24, AIME25, HumanEval, LiveCodeBench, and SWE-Bench Verified, despite having approximately one-third as many parameters.

\textbf{Token-efficient acquisition of complementary capabilities.} These results are achieved with only 1.2T pre-training tokens and fewer than 2T tokens overall. In comparison, SmolLM3-3B is pre-trained on 11.2T tokens ($9.3\times$ more), while the preceding Qwen3 family, including Qwen3-1.7B, reports approximately 36T pre-training tokens ($30\times$ more). Despite its limited token budget, \ours{} achieves the best or near-best performance (within 1.0 percentage point) among models with fewer than 3.5B total parameters on 12 of 14 benchmarks, and the best performance across all evaluated models on five. Together, these results highlight the potential of jointly designed reasoning and agentic training to develop complementary capabilities within both model capacity and training token constraints.

\section{Experimental Setup}
\label{sec:recipe}

To facilitate reproducibility, we detail the model architecture and training recipe, including the data composition and training configurations across pre-training, mid-training, and post-training.

\subsection{Model Architecture}
Our architecture builds upon the designs of MobileLLM~\citep{liu2024mobilellm} and Muse-Glimmer-30B~\citep{meta2026museglimmer}. We adopt the Muse-Glimmer tokenizer, which has a 202{,}048-token subword vocabulary, together with SwiGLU~\citep{shazeer2020glu} activations and RoPE~\citep{su2024roformer} with $\theta=10^7$. We share the input and output embedding weights. The complete architectural specifications are summarized in Table~\ref{tab:arch}.
\begin{table*}[h]
\small
\centering
\caption{\small Detailed architecture specifications of $\ours{}$.
"Dim" denotes the embedding dimension and "Hidden Dim" represents the dimension inside the feed-forward network.}
\label{tab:arch}
\vspace{-0.5em}
\setlength{\tabcolsep}{1.5mm}
\renewcommand{\arraystretch}{1.02}
\begin{tabular}{lrrrrrr}
\toprule
\textbf{Model} & \textbf{Layer} & \textbf{Head} & \textbf{KV-Head} & \textbf{Dim} & \textbf{Hidden Dim} & \textbf{Params} \\
\midrule
$\ours{}$-2.5B & 32 &  32 & 8 & 2048 & 8192 & 2.49B \\
\bottomrule
\end{tabular}
\end{table*}

\subsection{Training Recipe}
The complete set of training hyperparameters is detailed in Table~\ref{tab:train_recipe}, with stage-specific configurations as follows:

\textbf{Pre-training.}
The model is optimized with Muon at a peak learning rate of $1.2\times10^{-3}$ and a weight decay coefficient of $0.1$; its AdamW component uses $\beta_1=0.9$, $\beta_2=0.95$, and $\epsilon=10^{-12}$. We employ a cosine learning-rate schedule with $7{,}500$ warmup steps. Training further incorporates \emph{online} knowledge distillation from Muse-Glimmer-30B~\citep{meta2026museglimmer}, using pure forward KL divergence with no hard-label loss.

\textbf{Mid-training.}
Starting from the pre-trained checkpoint, we perform two consecutive 50,000-step knowledge-distillation phases. We distill from Muse-Glimmer-30B using pure forward KL divergence, with no hard-label loss. The student adopts the teacher's tokenizer, so the distillation loss covers the full 202{,}048-entry vocabulary, and both the teacher's and the student's output logits are soft-capped at 20. As in pre-training, we use Muon with a weight decay coefficient of $0.1$, with $\epsilon=10^{-8}$ for its AdamW component; in each phase the learning rate starts at $3.6279\times10^{-4}$ and decays linearly to zero without warmup, with 2M ($2^{21}$) tokens per step, or about 105B tokens per phase. The two phases draw from the same mid-training data sources shown in Table~\ref{table:mid-train-data} and differ only in their data mixtures. After normalization, the proportions of mathematics and code data increase from 15.1\% and 27.6\% in Phase~1 to 49.0\% and 39.2\% in Phase~2, respectively.

\textbf{Post-training.}
Post-training consists of three sequential stages. The first General SFT stage combines a 100K-example uniform sample of \href{https://huggingface.co/datasets/nvidia/Nemotron-Agentic-v1}{Nemotron-Agentic-v1} with the full \href{https://huggingface.co/datasets/allenai/Dolci-Think-SFT-7B}{Dolci-Think-SFT-7B} corpus and trains for two epochs. The second Agentic Reasoning SFT stage jointly optimizes multi-turn agentic trajectories and long-horizon mathematical, coding, and scientific reasoning data, taking one full pass over each reasoning source; agentic data makes up 16.0\% of the examples and 13.2\% of the supervised tokens. The agentic data includes fourteen passes over \href{https://huggingface.co/datasets/open-thoughts/OpenThoughts-Agent-SFT-100K}{OpenThoughts-Agent-SFT-100K}~\citep{raoof2026openthoughts} and three passes over 27k synthesized examples, while the reasoning data comprises the CoT split of \href{https://huggingface.co/datasets/nvidia/OpenMathReasoning}{OpenMathReasoning}, \href{https://huggingface.co/datasets/nvidia/OpenCodeReasoning-2}{OpenCodeReasoning-2}, and \href{https://huggingface.co/datasets/nvidia/OpenScienceReasoning-2}{OpenScienceReasoning-2}.
All sources are jointly mixed for SFT training. All SFT stages use a maximum sequence length of 32,768, RoPE $\theta=10^7$, no sequence packing, and loss masks stored over assistant tokens. Both stages use Muon with a global batch of 1{,}024 examples, a cosine learning-rate schedule with 10\% warmup, zero weight decay, and gradient clipping at 1.0; the peak learning rate is $8\times10^{-5}$ for General SFT and $4\times10^{-5}$ for Agentic Reasoning SFT, which trains for one epoch. The final on-policy knowledge distillation (OPD) stage minimizes the forward KL divergence from teacher to student over the full 202,048-token vocabulary. For each prompt, we sample one student rollout at temperature 0.6 and top-$p$ 0.95, with a maximum of 30,000 generated tokens. Since both models share the same tokenizer, the teacher directly scores the student's generated tokens at each position.

\begin{table}[t]
\small
\centering
\caption{\small Training setup for \ours{}.}
\vspace{-0.5em}
\label{tab:train_recipe}
\setlength{\tabcolsep}{1.5mm}
\renewcommand{\arraystretch}{1.02}
\begin{tabular}{@{}llrrrrrr@{}}
\toprule
\textbf{Stage} & \textbf{Phase} & \textbf{Tokens / Samples} & \textbf{Batch Size} & \textbf{Seq. Len.} & \textbf{Updates / Epochs} & \textbf{Peak LR} & \textbf{\#GPUs} \\
\midrule
\textit{Pre-training} & Single phase & 1.2T tokens & 128 & 16k & 600k & 1.2e-3 & 32$\times$8 \\
\midrule
\multirow{2}{*}{\textit{Mid-training}}
 & Phase 1 & 105B tokens & 128 & 16k & 50k & 3.63e-4 & $16{\times}8$ \\
 & Phase 2 & 105B tokens & 128 & 16k & 50k & 3.63e-4 & $16{\times}8$ \\
\midrule
\multirow{3}{*}{\textit{Post-training}}
 & General SFT & 2.4M samples & 1024 & 32k &  2 epochs & 8e-5 & $16{\times}8$ \\
 & Agentic Reasoning SFT & 8.8M samples & 1024 & 32k & 1 epoch & 4e-5 & $16{\times}8$ \\
 & On-policy KD & 812K samples & 1024 & 32k & 1 epoch & 5e-6 & $16{\times}8$ \\

\bottomrule
\end{tabular}
\end{table}

\section{Related Work}
\label{sec:related_work}

Small language models have improved through advances in architecture design, data curation, knowledge distillation, and multi-stage training. MobileLLM established strong sub-billion-parameter architectures and demonstrated that compact models can support both conversational and API-calling workloads~\citep{liu2024mobilellm}. Subsequent models, including SmolLM3 and Qwen3, have further improved small-model capabilities through curated and synthetic data, extended pre-training, distillation, and increasingly sophisticated post-training pipelines~\citep{bakouch2025smollm3,yang2025qwen3}. However, these approaches often rely on large training budgets: SmolLM3 and Qwen3 use approximately 11.2T and 36T pre-training tokens, respectively.

Toward more accessible and reproducible training recipes, MobileLLM-R1 is particularly close to our setting, providing an open pre-training-to-post-training recipe for a sub-billion-parameter model using 4T pre-training tokens, but focuses on mathematical and coding reasoning rather than agentic tool use~\citep{zhao2026mobilellm}. Qwen3.5-2B/4B provides a strong small-model agentic baseline, but does not release a fully reproducible end-to-end recipe, including key details of the training data composition and token budgets~\citep{team2026qwen3}. OLMo3 is the closest precedent in terms of openness: it releases the complete model development flow and incorporates function calling and multi-step tool-use trajectories, including interactions with real MCP environments~\citep{olmo2025olmo}. Its smallest model, however, is 7B parameters, with approximately 5.9T pre-training tokens. In comparison, 2B-scale models are better suited to on-device deployment under tight memory, compute, and latency constraints~\citep{yan2025we}. Thus, despite these advances, open end-to-end recipes at this scale remain relatively unexplored.

Beyond model-level recipes, prior work has also explored data-efficient training through skill-aware data selection and regression-based mixture optimization~\citep{chen2023skill, lauffer2025imitation, brown2024large, liu2407regmix}. Our work complements these directions by examining how data composition and training decisions across pre-training, mid-training, and post-training jointly affect reasoning and agentic capabilities in compact models under constrained training budgets.

\section{Conclusion}
We show that building capable small agentic models requires more than effective post-training: agentic trainability is strongly shaped by the pre-training trajectory. Our findings highlight three key principles: pre-training should preserve diverse behavioral support, long-horizon deliberation and multi-turn agentic reasoning exhibit distinct behavioral patterns yet can be effectively learned jointly, and effective on-policy distillation requires both structural and trajectory distribution alignment established during SFT.
Together, these results suggest that pre-training and post-training should be co-designed as a unified process. With only 1.2T pre-training tokens, our 2.5B model achieves strong performance across reasoning, tool use, and software engineering, demonstrating that capable agentic models can be developed without frontier-scale training budgets. We release the complete training recipe and checkpoints to facilitate further research on efficient and reproducible agentic model training.

\clearpage
\newpage
\bibliographystyle{assets/plainnat}
\bibliography{paper}

\clearpage
\appendix

\section{Ablation Study of Cross-Domain Bridge Data}
\label{app:bridge_ablation}

To quantify the contribution of cross-domain bridge data, we conduct a controlled ablation comparing training with and without the bridge-data component in pre-training. Both variants are evaluated at 50K training steps using the same evaluation protocol across 19 downstream benchmarks.

\begin{table}[t]
\small
\centering
\caption{\small{Ablation of cross-domain bridge data at 50K training steps.  $\Delta_{\mathrm{abs}}$ denotes the absolute accuracy difference in percentage points (pp), while $\Delta_{\mathrm{rel}}$ denotes the relative improvement over the no-bridge baseline. The mean is computed with equal weight across all 19 benchmarks. Incorporating bridge data improves performance on all evaluated benchmarks.}}
\label{tab:bridge_ablation}
\renewcommand{\arraystretch}{1.02}
\setlength{\tabcolsep}{1.5mm}
\vspace{-0.5em}
\begin{tabular}{lrrrr}
\toprule
\textbf{Benchmark}
& \textbf{w/o Bridge}
& \textbf{w/ Bridge}
& $\boldsymbol{\Delta_{\mathrm{abs}}}$
& $\boldsymbol{\Delta_{\mathrm{rel}}}$ \\
\midrule
MMLU          & 36.88 & 38.47 & +1.59 & +4.31\% \\
MMLU-Redux    & 38.97 & 40.30 & +1.33 & +3.41\% \\
MMLU-Pro      & 15.92 & 16.53 & +0.61 & +3.83\% \\
ARC-Challenge & 43.90 & 45.15 & +1.25 & +2.85\% \\
ARC-Easy      & 67.69 & 70.66 & +2.97 & +4.39\% \\
HellaSwag     & 63.97 & 66.30 & +2.33 & +3.64\% \\
PIQA          & 74.28 & 76.22 & +1.94 & +2.61\% \\
SIQA          & 46.02 & 47.39 & +1.37 & +2.98\% \\
WinoGrande    & 62.96 & 64.48 & +1.52 & +2.41\% \\
BoolQ         & 61.75 & 63.15 & +1.40 & +2.27\% \\
GSM8K         & 45.94 & 47.99 & +2.05 & +4.46\% \\
MATH          & 15.70 & 16.40 & +0.70 & +4.46\% \\
BBH           & 24.26 & 25.07 & +0.81 & +3.34\% \\
BBEH          &  3.34 &  3.43 & +0.09 & +2.69\% \\
GPQA          & 13.02 & 13.48 & +0.46 & +3.53\% \\
SuperGPQA     & 10.06 & 10.45 & +0.39 & +3.88\% \\
HumanEval+    & 26.73 & 27.44 & +0.71 & +2.66\% \\
MBPP          & 42.96 & 44.44 & +1.48 & +3.45\% \\
CRUXEval      & 23.12 & 24.25 & +1.13 & +4.89\% \\
\midrule
\textbf{Mean}
& 37.76 & 39.03
& \textbf{+1.27}
& \textbf{+3.48\%} \\
\bottomrule
\end{tabular}
\end{table}

As shown in Table~\ref{tab:bridge_ablation}, incorporating cross-domain bridge data improves performance across all 19 evaluated benchmarks. The average absolute improvement is 1.27 percentage points, corresponding to a 3.48\% macro-averaged relative improvement. The largest absolute gains are observed on ARC-Easy (+2.97 pp), HellaSwag (+2.33 pp), and GSM8K (+2.05 pp).

\section{Memory Footprint and Inference Efficiency}
\label{app:on_device_profiling}

To assess the feasibility of deploying $\ours{}$ on memory-constrained devices, we profile the memory footprint and inference efficiency of four models using \texttt{llama.cpp} on an Apple M4 Max laptop. All models are evaluated with Q4\_K\_M GGUF weight quantization, an 8-bit KV cache, and a single concurrent request. Table~\ref{tab:on_device_profiling} reports the peak memory footprint at a 32K-token context length, along with time to first token (TTFT) and decoding throughput at 4K and 32K prompt lengths.

\begin{table*}[t]
\small
\centering
\caption{\small Memory footprint and inference efficiency measured using \texttt{llama.cpp} on an Apple M4 Max laptop. All models use Q4\_K\_M weight quantization and an 8-bit KV cache, with a single concurrent request.}
\vspace{-0.5em}
\label{tab:on_device_profiling}
\renewcommand{\arraystretch}{1.02}
\setlength{\tabcolsep}{1.5mm}
\begin{tabular}{lcccc}
\toprule
\textbf{Model} &
\textbf{Memory (32K context)} &
\textbf{Within 4 GiB budget} &
\textbf{TTFT (4K / 32K)} &
\textbf{Decode tokens/s (4K / 32K)} \\
\midrule
$\ours{}$ (2.5B) & 3.0 GiB & Yes & 1.7 s / 32 s & 137 / 84 \\
Qwen3.5-2B   & 2.0 GiB & Yes & 1.1 s / 11 s & 190 / 159 \\
Qwen3.5-4B   & 3.8 GiB & Yes & 2.7 s / 31 s & 101 / 81 \\
Qwen3.5-9B   & 6.1 GiB & No  & 5.1 s / 52 s & 66 / 56 \\
\bottomrule
\end{tabular}
\end{table*}

At a 32K-token context length, $\ours{}$ requires 3.0 GiB of memory, comprising approximately 1.5 GiB for model weights, 1.1 GiB for the KV cache, and 0.4 GiB for compute buffers. This footprint falls within an approximate 4 GiB per-app memory budget on an 8 GB iPhone, suggesting the feasibility of memory-constrained deployment. Under the same budget, models up to approximately 4B parameters remain within the measured memory limit, consistent with the parameter scale of Apple's approximately 3B-parameter on-device model~\citep{li2025apple}. However, actual deployment feasibility depends on device-specific memory limits, runtime overhead, and hardware characteristics, which are not captured by our laptop-based measurements.

The reported memory footprints also suggest that all four models could potentially be accommodated within the physical memory capacity of an 8 GB laptop, such as an M1 MacBook Air, although available memory and runtime overhead may impose additional constraints. Overall, these results provide an initial characterization of the memory requirements and inference efficiency of $\ours{}$ under a standardized quantized inference configuration.

\section{Results Across Training Stages}
\label{sec:post_stages}
\begin{table}[t]
\small
\centering
\caption{\small Evaluation across pre-training and mid-training checkpoints. WikiText-2 reports perplexity (lower is better), while all other benchmarks report accuracy (higher is better).}
\label{tab:base_ckpts}
\vspace{-0.5em}
\setlength{\tabcolsep}{1.5mm}
\renewcommand{\arraystretch}{1.02}
\begin{tabular}{@{}lccccc@{}}
\toprule
\textbf{Benchmark} & \multicolumn{3}{c}{\textbf{Pre-training}} & \multicolumn{2}{c}{\textbf{Mid-training}} \\
\cmidrule(lr){2-4}\cmidrule(lr){5-6}
 & \textbf{200K} & \textbf{400K} & \textbf{600K} & \textbf{Phase 1} & \textbf{Phase 2} \\
\midrule
MATH500 ($\uparrow$)    & 22.4 & 28.4 & 31.6 & 37.2 & 45.0 \\
GSM8K ($\uparrow$)       & 62.4 & 69.0 & 77.0 & 86.3 & 88.9 \\
MBPP ($\uparrow$)        & 36.8 & 38.2 & 44.2 & 52.2 & 54.2 \\
HumanEval ($\uparrow$)   & 36.6 & 42.1 & 51.2 & 43.3 & 43.9 \\
CommonSense ($\uparrow$) & 58.0 & 59.8 & 61.0 & 60.7 & 56.7 \\
MMLU ($\uparrow$)        & 48.1 & 52.6 & 58.9 & 61.0 & 54.4 \\
WikiText-2 (Perplexity $\downarrow$)  & 16.12 & 15.41 & 14.74 & 12.22 & 15.75 \\
\bottomrule
\end{tabular}
\end{table}

\begin{table*}[t]
\small
\centering
\caption{\small Impact of starting checkpoints on downstream post-training performance. The same SFT recipe is applied to checkpoints from different pre-training and mid-training stages, without OPD. All metrics report accuracy.}
\label{tab:recipe_by_start}
\vspace{-0.5em}
\setlength{\tabcolsep}{1.5mm}
\renewcommand{\arraystretch}{1.02}
\begin{tabular}{@{}lcccccccccc@{}}
\toprule
& \multicolumn{5}{c}{\textbf{General SFT}}
& \multicolumn{5}{c}{\textbf{Agentic Reasoning SFT}} \\
\cmidrule(lr){2-6}\cmidrule(lr){7-11}
\textbf{Benchmark}
& \multicolumn{3}{c}{\textbf{Pre-training}}
& \multicolumn{2}{c}{\textbf{Mid-training}}
& \multicolumn{3}{c}{\textbf{Pre-training}}
& \multicolumn{2}{c}{\textbf{Mid-training}} \\
\cmidrule(lr){2-4}\cmidrule(lr){5-6}
\cmidrule(lr){7-9}\cmidrule(lr){10-11}
& \textbf{200K} & \textbf{400K} & \textbf{600K}
& \textbf{Phase 1} & \textbf{Phase 2}
& \textbf{200K} & \textbf{400K} & \textbf{600K}
& \textbf{Phase 1} & \textbf{Phase 2} \\
\midrule
AIME24 ($\uparrow$)
& 25.0 & 32.5 & 40.8 & 42.5 & 50.0
& 29.6 & 36.3 & 36.5 & 45.0 & 49.8 \\
AIME25 ($\uparrow$)
& 29.2 & 35.0 & 32.5 & 36.7 & 41.5
& 26.5 & 27.3 & 32.5 & 36.3 & 39.0 \\
MATH500 ($\uparrow$)
& 77.4 & 81.4 & 83.4 & 86.2 & 88.6
& 85.4 & 84.8 & 87.6 & 90.2 & 90.2 \\
LiveCodeBench ($\uparrow$)
& 11.5 & 15.1 & 17.9 & 21.3 & 22.8
& 21.9 & 26.8 & 27.9 & 32.6 & 38.6 \\
GPQA ($\uparrow$)
& 30.2 & 32.7 & 36.2 & 33.7 & 33.3
& 29.2 & 32.1 & 35.6 & 41.5 & 35.7 \\
BFCL V4 ($\uparrow$)
& 29.6 & 32.7 & 34.0 & 37.0 & 38.9
& 39.7 & 41.1 & 46.8 & 47.9 & 48.1 \\
\bottomrule
\end{tabular}
\end{table*}

\begin{table}[t]
\small
\centering
\caption{\small Performance across post-training stages. Stage 1 and Stage 2 denote General SFT and Agentic Reasoning SFT, respectively, followed by on-policy distillation (OPD). }
\label{tab:post_stages}
\vspace{-0.5em}
\setlength{\tabcolsep}{1.5mm}
\renewcommand{\arraystretch}{1.02}
\begin{tabular}{@{}llccc@{}}
\toprule
\textbf{Category} & \textbf{Benchmark} & Stage1 SFT & Stage2 SFT & OPD \\
\midrule
\multirow{4}{*}{Math}
  & AIME24 & 50.0 & 49.8 & 59.4 \\
  & AIME25 & 41.5 & 39.0 & 49.0 \\
  & MATH500 & 88.6 & 90.2 & 91.0 \\
  & GSM8K & 90.0 & 93.0 & 91.9 \\
\midrule
\multirow{2}{*}{Knowledge}
  & MMLU & 52.7 & 55.9 & 56.5 \\
  & GPQA & 33.3 & 35.7 & 45.2 \\
\midrule
\multirow{3}{*}{Coding}
  & HumanEval & 55.5 & 64.0 & 69.5 \\
  & MBPP & 50.6 & 50.8 & 49.6 \\
  & LiveCodeBench & 22.8 & 38.6 & 47.3 \\
\midrule
\multirow{2}{*}{Agentic}
  & BFCL V4 & 38.9 & 48.1 & 50.6 \\
  & SWE-Bench Verified & -- & 25.0 & 30.9 \\
\midrule
\multirow{3}{*}{OOD}
  & Terminal-Bench & -- & 4.1 & 5.2 \\
  & Aider Polyglot & -- & 4.3 & 6.5 \\
  & BFCL-Parity & 82.7 & 75.6 & 77.5 \\
\midrule
Instruction following & IFEval & 68.6 & 52.5 & 67.1 \\
\bottomrule
\end{tabular}
\end{table}

\begin{table}[!htb]
\small
\centering
\caption{\small BFCL V4 accuracy across tool-use and interaction categories. Stage 2 SFT denotes the checkpoint after Agentic Reasoning SFT, while OPD denotes the checkpoint after subsequent on-policy distillation.}
\label{tab:bfcl_breakdown}
\vspace{-0.5em}
\begingroup
\setlength{\tabcolsep}{1.5mm}
\renewcommand{\arraystretch}{1.02}
\begin{tabular}{@{}lcc@{}}
\toprule
\textbf{Category} & Stage2 SFT & OPD \\
\midrule
Non-live & 58.3 & 60.9 \\
\quad Simple & 60.8 & 61.0 \\
\quad Multiple & 88.0 & 90.0 \\
\quad Parallel & 45.5 & 47.0 \\
\quad Parallel multiple & 39.0 & 45.5 \\
\midrule
Live & 71.8 & 73.2 \\
\quad Simple & 75.2 & 74.8 \\
\quad Multiple & 72.3 & 74.2 \\
\quad Parallel & 43.8 & 50.0 \\
\quad Parallel multiple & 33.3 & 29.2 \\
\midrule
Irrelevance & 84.8 & 82.5 \\
Relevance & 87.5 & 68.8 \\
\midrule
Multi-turn & 38.6 & 43.5 \\
\quad Base & 52.0 & 60.0 \\
\quad Missing function & 27.0 & 40.0 \\
\quad Missing parameter & 38.5 & 37.5 \\
\quad Long context & 37.0 & 36.5 \\
\midrule
Web search & 67.5 & 62.0 \\
\quad Base & 77.0 & 70.0 \\
\quad No snippet & 58.0 & 54.0 \\
\midrule
Memory & 7.5 & 17.4 \\
\quad Key-value & 11.6 & 13.6 \\
\quad Vector & 5.2 & 21.9 \\
\quad Recursive summary & 5.8 & 16.8 \\
\midrule
Overall & 48.1 & 50.6 \\
\bottomrule
\end{tabular}
\endgroup
\end{table}

\paragraph{Pre-training and mid-training progressively shape model capabilities.}
As shown in Table~\ref{tab:base_ckpts}, continued pre-training improves all benchmarks, with MATH500 increasing from 22.4 to 31.6 and WikiText-2 perplexity decreasing from 16.12 to 14.74. Mid-training further strengthens mathematical reasoning, raising MATH500 to 45.0 and GSM8K to 88.9. However, these gains are not uniform: Phase 1 improves most metrics but reduces HumanEval accuracy, while Phase 2 further improves math and MBPP at the expense of general knowledge, commonsense reasoning, and language modeling perplexity. These results highlight a trade-off between targeted reasoning specialization and the retention of broader capabilities.

\paragraph{Pre-training gains persist through post-training.}
To isolate the contribution of the starting checkpoint, we apply identical SFT stages to five checkpoints without OPD. As shown in Table~\ref{tab:recipe_by_start}, later checkpoints generally yield stronger post-trained models across mathematical reasoning, coding, and tool use. These results reinforce that the effectiveness of post-training is fundamentally conditioned on the capability foundation established during pre-training: post-training can amplify existing capabilities but does not readily recover those underdeveloped or compromised during pre-training.

\paragraph{Agentic Reasoning SFT and OPD provide complementary gains across reasoning and interaction tasks.} As shown in Table~\ref{tab:post_stages}, Agentic Reasoning SFT improves BFCL V4 from 38.9 to 48.1 and GSM8K from 90.0 to 93.0, but slightly reduces AIME performance. Subsequent OPD substantially improves long-horizon reasoning, raising AIME'24/25 by 9.6/10.0 points, LiveCodeBench by 8.7 points, and 5.9 on SWE-Bench Verified. These gains extend beyond the distillation domains, with improvements of 9.5 points on GPQA and 14.6 on IFEval. The broad improvements across reasoning, agentic, and out-of-distribution benchmarks suggest that OPD transfers capabilities beyond the specific tasks used for distillation, rather than merely improving performance on the training domains.

\paragraph{BFCL: OPD Improves Multi-Turn Tool Use}
Table~\ref{tab:bfcl_breakdown} breaks BFCL V4 down by category. OPD yields more pronounced gains in multi-turn interactions than in single-turn function calling. Unlike single-turn tasks, which can be learned from static supervised examples, multi-turn interactions require the model to operate under its own evolving trajectory distribution.

\section{Teacher Intervention Strategies for Agentic Behavior}
\label{sec:opd_agentic}

Supervised fine-tuning learns from expert trajectories, whereas deployment exposes the model to its own prefixes, where an early tool-selection error can induce an unsupported state and compound throughout the episode. On-policy knowledge distillation (OPD)~\citep{agarwal2024policy} addresses this mismatch by supervising states reached by the student itself, but tool use introduces a causal asymmetry: once an action produces an invalid observation, subsequent token-level guidance cannot recover the counterfactual trajectory. This motivates controlling \emph{where} the teacher intervenes and \emph{how long} it remains responsible for the correction.

We study a progression from token-level guidance to trajectory-level correction. \textit{Naive OPD}~\citep{agarwal2024policy} applies dense token-level supervision via KL divergence at every student-generated position, but continues training on potentially corrupted prefixes. \textit{One-Turn Supervision} instead uses a teacher-generated next-turn response as a counterfactual target at each student-visited state, without executing the teacher response or backpropagating loss through the preceding student action~\citep{sun2026step,li2026revisiting,li2026policy}. \textit{Till-End Supervision} extends this correction by letting the teacher complete the trajectory from a sampled student state and supervising the entire suffix, providing recovery beyond a single turn~\citep{lauffer2025imitation}. Finally, \textit{Verified Till-End} retains only structurally complete and outcome-correct teacher suffixes, trading coverage for higher-confidence trajectory-level supervision.

We conduct controlled experiments to compare the effect of intervention granularity, using the same 12,180 candidate episodes and rollout horizon. As shown in Table~\ref{tab:teacher_supervision_ablation}, \textit{One-Turn Supervision} improves BFCL multi-turn accuracy from 19.00\% with \textit{Naive OPD} to 24.88\%, indicating that localized counterfactual correction is more effective than dense token-level supervision on student-corrupted prefixes. \textit{Till-End} further reaches 30.75\% accuracy by extending supervision through the observation-conditioned recovery trajectory. Consistent with this, analysis of raw evaluation traces shows that tool-execution errors drop from 7.63\% to 0.99\% of tool-executing steps, while failed-action repetition drops from 74\% to 8\%. As shown in Table~\ref{tab:teacher_supervision_length}, \textit{Till-End} also provides denser supervision, with a loss-to-input token ratio of 27.83\%, compared with 6.51\% for \textit{One-Turn Supervision}. \textit{Verified Till-End}, however, falls to 16.50\% in multi-turn accuracy, mainly because only 37.2\% of switch-reached suffixes pass verification, filtering out difficult recovery states and reducing coverage. Accordingly, it retains only 4,082 episodes from the same 12,180 candidates, reducing loss tokens from 14.77M to 3.91M.

\begin{table*}[t]
\footnotesize
\centering
\caption{\small BFCL V4 multi-turn accuracy comparison on teacher supervision design choices. }
\vspace{-0.5em}
\label{tab:teacher_supervision_ablation}
\setlength{\tabcolsep}{1.2mm}
\renewcommand{\arraystretch}{1.02}
\begin{tabular}{@{}lccccccc@{}}
\toprule
\textbf{Model / recipe} & \textbf{Optimization Objective} & \textbf{MT Avg.} & \textbf{$\Delta$} &
\textbf{Base} & \textbf{Miss Func.} & \textbf{Miss Param.} &
\textbf{Long Ctx.} \\
\midrule
Stage1 SFT initialization & -- & 14.25 & 0.00 & 17.00 & 16.00 & 16.50 & 7.50 \\
\midrule
Exp.~1: Naive OPD & KLD loss & 19.00 & +4.75 & 33.50 & 11.50 & 19.50 & 11.50 \\
Exp.~2: One-Turn Supervision & CE loss & 24.88 & +10.63 & 38.50 & \textbf{19.50} & 25.50 & 16.00 \\
Exp.~3: Till-End Supervision & CE loss & \textbf{30.75} & \textbf{+16.50} & \textbf{50.50} & 17.00 & \textbf{34.50} & \textbf{21.00} \\
Exp.~4: Verified Till-End Supervision & CE loss & 16.50 & +2.25 & 29.00 & 6.00 & 19.00 & 12.00 \\
\bottomrule
\end{tabular}
\end{table*}

\begin{table*}[t]
\footnotesize
\centering
\caption{\small Training corpus statistics across teacher intervention strategies. Candidate episodes counts sampled rollout episodes before row construction and filtering. Training rows counts serialized examples consumed by the optimizer after expansion and filtering. Supervised turns counts assistant messages whose token spans carry nonzero loss. Input tokens counts all serialized tokens processed by the optimizer, including masked context, whereas loss tokens counts only tokens with nonzero loss. Loss/input is their ratio.}
\vspace{-0.5em}
\label{tab:teacher_supervision_length}
\setlength{\tabcolsep}{1.5mm}
\renewcommand{\arraystretch}{1.02}
\begin{tabular}{@{}lrrrrrrr@{}}
\toprule
Recipe & Candidate eps. & Training rows & Sup. turns & Input tok. & Loss tok. & Loss/input \\
\midrule
Exp.~1: Naive OPD & 12{,}180 & 12{,}180 & 67{,}265 & 70.58M & 35.04M & 49.65\% \\
Exp.~2: One-Turn Supervision & 12{,}180 & 59{,}377 & 59{,}377 & 300.91M & 19.60M & 6.51\% \\
Exp.~3: Till-End Supervision & 12{,}180 & 10{,}967 & 70{,}431 & 53.08M & 14.77M & 27.83\% \\
Exp.~4: Verified Till-End Supervision & 12{,}180 & 4{,}082 & 26{,}170 & 15.60M & 3.91M & 25.08\% \\
\bottomrule
\end{tabular}
\end{table*}

\section{OPD Rollout Generation Details}

On-policy distillation (OPD) trains the student on its own generated trajectories, or rollouts, with the teacher providing token-level supervision. These rollouts expose the teacher to the student's actual generation distribution, making their quality and interaction structure critical to effective distillation.

Rollouts that reach the length cap usually end in a repetition loop; we cut them 64 tokens after the detected loop start and keep them, because the loop start is where the two models disagree most. Capped rollouts without a detectable loop and malformed rollouts are also dropped.
Code requires explicit channel alignment. We observe that 95\% of code rollouts from \ourssft{} bypass the reasoning channel and directly generate user-visible responses. To encourage reasoning before code generation while aligning student trajectories with the teacher's behavioral distribution, we explicitly enforce the \cttok{to=self} header during rollout generation for both coding and agentic OPD. This ensures that trajectories begin with internal reasoning before proceeding to code generation or tool calls. Importantly, the resulting OPD model learns to maintain this reasoning-first interaction pattern without requiring explicit channel enforcement at inference time.

\section{Evaluation Protocol}
We evaluate all models under a unified inference and evaluation pipeline spanning mathematical reasoning, knowledge, coding, tool use, and agentic tasks. Unless otherwise specified, evaluations use zero-shot prompting with the model's native chat template and enable thinking for reasoning-oriented benchmarks. All generative evaluations use a standardized decoding configuration of temperature $0.6$, top-$p$ $0.95$, and top-$k$ $20$, while benchmarks with deterministic evaluation use greedy decoding. We report the benchmark-native metric over the full evaluation set.

For mathematical reasoning, we evaluate on AIME'24~\citep{jia2024aime}, AIME'25~\citep{matharena_aime2025}, MATH500 \citep{hendrycks2021math}, and GSM8K~\citep{cobbe2021training}. AIME is evaluated with 16 sampled generations per problem and reported using the corresponding average accuracy across samples, while MATH500 and GSM8K use single-sample pass@1. We use up to 30K generated tokens for long-horizon reasoning, subject to each model's supported context length. For knowledge evaluation, we use 5-shot MMLU~\citep{hendrycks2020mmlu} with log-probability scoring and report the macro-average across its four subject groups. GPQA Diamond~\citep{rein2023gpqa} is evaluated zero-shot with shuffled answer options and four sampled generations per question, and we report accuracy averaged over the four samples.

For code generation, we evaluate HumanEval~\citep{chen2021evaluating} and MBPP~\citep{austin2021program}, using pass@1, with 0-shot and 3-shot prompting, respectively. We additionally evaluate LiveCodeBench (v6)~\citep{jain2025livecodebench} using 16 samples per problem and report the mean pass rate over all graded samples. For agentic tool use, we evaluate BFCL v4~\citep{patil2025berkeley} using its full suite, and report the benchmark's overall accuracy. SWE-Bench Verified~\citep{jimenez2024swe} is evaluated through a standardized terminal-agent scaffold with 4 independent attempts per task, and we report task-level mean pass@1.

We further assess out-of-distribution agentic capabilities on Terminal-Bench 2~\citep{merrill2026terminal}, Aider Polyglot~\citep{aider2024polyglot}, and BFCL-Parity. Each task is evaluated with three independent attempts under the same terminal-agent scaffold, and we report the mean success rate over all graded task attempts. BFCL-Parity is evaluated as a separate 123-task agentic benchmark with a standardized Terminus-based scaffold and task construction distinct from native BFCL V4, and is therefore reported independently rather than as a BFCL V4 result.

Unless otherwise noted, agentic evaluations use a maximum of 4K generated tokens per turn and a 32K input-context budget, with automatic context summarization for long trajectories. All evaluations are conducted using the same inference stack and model-specific supported context limits, ensuring a consistent evaluation protocol across model families.

\FloatBarrier

\end{document}